\documentclass{article}

\usepackage{amsmath,amsfonts,bm}

\def\eqref#1{equation~\ref{#1}}

\def\1{\bm{1}}

\def\vp{{\bm{p}}}

\DeclareMathAlphabet{\mathsfit}{\encodingdefault}{\sfdefault}{m}{sl}
\SetMathAlphabet{\mathsfit}{bold}{\encodingdefault}{\sfdefault}{bx}{n}

\newcommand{\cpuct}{c_{\mathrm{puct}}}

\def\vp{\mathbf{p}}
\def\vpi{\boldsymbol{\pi}}
\usepackage{amsmath}
\usepackage{amssymb}
\usepackage{booktabs}
\usepackage{caption}
\usepackage{changepage}
\usepackage{chngcntr}
\usepackage{comment}
\usepackage{csquotes}
\usepackage{draftcopy}
\usepackage{graphicx}
\usepackage{makecell}
\usepackage{mathtools}
\usepackage{microtype}
\usepackage{subcaption}
\usepackage{tcolorbox}
\usepackage{thm-restate}
\usepackage{thmtools}
\usepackage{threeparttable}
\usepackage{url}
\usepackage{verbatim}

\usepackage[hidelinks]{hyperref}
\usepackage[capitalise,nameinlink]{cleveref}

\def\<{\langle}
\def\>{\rangle}

\DeclarePairedDelimiter\abs{\lvert}{\rvert}
\makeatletter
\let\oldabs\abs
\def\abs{\@ifstar{\oldabs}{\oldabs*}}
\makeatother

\crefname{equation}{}{}

\DeclareRobustCommand{\elfopengo}{ELF OpenGo}
\DeclareRobustCommand{\system}{\texttt{ELF$^{++}$}}

\usepackage[inline,nomargin,index]{fixme}
\FXRegisterAuthor{yt}{ayt}{\color{red}Yuandong}
\FXRegisterAuthor{jm}{ajm}{\color{blue}Jerry}
\FXRegisterAuthor{qg}{aqg}{\color{green}Qucheng}
\FXRegisterAuthor{ss}{ass}{\color{magenta}Shubho}
\FXRegisterAuthor{zc}{azc}{\color{orange}Zhuoyuan}
\FXRegisterAuthor{jp}{ajp}{\color{teal}James}
\FXRegisterAuthor{lz}{alz}{\color{lightgray}Larry}

\usepackage{hyperref}

\usepackage[accepted]{icml2019}

\icmltitlerunning{\elfopengo{}: An Analysis and Open Reimplementation of AlphaZero}

\begin{document}

\twocolumn[
\icmltitle{\elfopengo{}: An Analysis and Open Reimplementation of AlphaZero}



\icmlsetsymbol{equal}{*}

\begin{icmlauthorlist}
\icmlauthor{Yuandong Tian}{fair}
\icmlauthor{Jerry Ma}{equal,fair}
\icmlauthor{Qucheng Gong}{equal,fair}
\icmlauthor{Shubho Sengupta}{equal,fair}
\icmlauthor{Zhuoyuan Chen}{fair}
\icmlauthor{James Pinkerton}{fair}
\icmlauthor{C. Lawrence Zitnick}{fair}
\end{icmlauthorlist}

\icmlaffiliation{fair}{Facebook AI Research, Menlo Park, California, USA}

\icmlcorrespondingauthor{Yuandong Tian}{yuandong@fb.com}
\icmlcorrespondingauthor{Jerry Ma}{maj@fb.com}
\icmlcorrespondingauthor{Larry Zitnick}{zitnick@fb.com}

\icmlkeywords{Machine Learning, ICML}

\vskip 0.3in
]



\printAffiliationsAndNotice{\icmlEqualContribution} 

\begin{abstract}

The AlphaGo, AlphaGo Zero, and AlphaZero series of algorithms are remarkable demonstrations of deep reinforcement learning's capabilities, achieving superhuman performance in the complex game of Go with progressively increasing autonomy. However, many obstacles remain in the understanding of and usability of these promising approaches by the research community. Toward elucidating unresolved mysteries and facilitating future research, we propose \elfopengo{}, an open-source reimplementation of the AlphaZero algorithm. \elfopengo{} is the first open-source Go AI to convincingly demonstrate superhuman performance with a perfect (20:0) record against global top professionals. We apply \elfopengo{} to conduct extensive ablation studies, and to identify and analyze numerous interesting phenomena in both the model training and in the gameplay inference procedures. Our code, models, selfplay datasets, and auxiliary data are publicly available.~\footnote{Resources available at \url{https://ai.facebook.com/tools/elf-opengo/}. Additionally, the supplementary appendix for this paper is available at \url{https://arxiv.org/pdf/1902.04522.pdf}.}

\end{abstract}

\section{Introduction}

The game of Go has a storied history spanning over 4,000 years and is viewed as one of the most complex turn-based board games with complete information. The emergence of AlphaGo~\citep{silver2016alphago} and its descendants AlphaGo Zero~\citep{silver2017alphagozero} and AlphaZero ~\citep{silver2018alphazero} demonstrated the remarkable result that deep reinforcement learning (deep RL) can achieve superhuman performance even without supervision on human gameplay datasets.

However, these advances in playing ability come at significant computational expense. A single training run requires millions of selfplay games and days of training on thousands of TPUs, which is an unattainable level of compute for the majority of the research community. When combined with the unavailability of code and models, the result is that the approach is very difficult, if not impossible, to reproduce, study, improve upon, and extend.

In this paper, we propose \elfopengo{}, an open-source reimplementation of the AlphaZero~\citep{silver2018alphazero} algorithm for the game of Go. We then apply \elfopengo{} toward the following three additional contributions.

First, we train a superhuman model for \elfopengo{}. After running our AlphaZero-style training software on 2,000 GPUs for 9 days, our 20-block model has achieved superhuman performance that is arguably comparable to the 20-block models described in \citet{silver2017alphagozero} and \citet{silver2018alphazero}. To aid research in this area we provide pre-trained superhuman models, code used to train the models, a comprehensive training trajectory dataset featuring 20 million selfplay games over 1.5 million training minibatches, and auxiliary data.~\footnote{Auxiliary data comprises a test suite for difficult ``ladder'' game scenarios, comparative selfplay datasets, and performance validation match logs (both vs. humans and vs. other Go AIs).} We describe the system and software design in depth and we relate many practical lessons learned from developing and training our model, in the hope that the community can better understand many of the considerations for large-scale deep RL.

Second, we provide analyses of the model's behavior during training. \textbf{(1)} As training progresses, we observe high variance in the model's strength when compared to other models. This property holds even if the learning rates are reduced. \textbf{(2)} Moves that require significant lookahead to determine whether they should be played, such as ``ladder'' moves, are learned slowly by the model and are never fully mastered. \textbf{(3)} We explore how quickly the model learns high-quality moves at different stages of the game. In contrast to tabular RL's typical behavior, the rate of progression for learning both mid-game and end-game moves is nearly identical in training \elfopengo{}.

Third, we perform extensive ablation experiments to study the properties of AlphaZero-style algorithms. We identify several important parameters that were left ambiguous in \citet{silver2018alphazero} and provide insight into their roles in successful training. We briefly compare the AlphaGo Zero and AlphaZero training processes. Finally, we find that even for the final model, doubling the rollouts in gameplay still boosts its strength by $\approx$ 200 ELO~\footnote{\citet{elo}'s method is a commonly-used performance rating system in competitive game communities.}, indicating that the strength of the AI is constrained by the model capacity. 

Our ultimate goal is to provide the resources and the exploratory insights necessary to allow both the AI research and Go communities to study, improve upon, and test against these promising state-of-the-art approaches.

\section{Related work}

In this section, we briefly review early work in AI for Go. We then describe the AlphaGo Zero~\citep{silver2017alphagozero} and AlphaZero~\citep{silver2018alphazero} algorithms, and the various contemporary attempts to reproduce them.

\paragraph{The game of Go} Go is a two-player board game traditionally played on a 19-by-19 square grid. The players alternate turns, with the black player taking the first turn and the white player taking the second. Each turn, the player places a stone on the board. A player can capture groups of enemy stones by occupying adjacent locations, or ``liberties''. Players can choose to ``pass'' their turn, and the game ends upon consecutive passes or resignation. In our setting, the game is scored at the end using Chinese rules, which award players one point for each position occupied or surrounded by the player. Traditionally, a bonus (``komi'') is given to white as compensation for going second. The higher score wins, and komi is typically a half-integer (most commonly $7.5$) in order to avoid ties.

\subsection{Early work}
\paragraph{Classical search}

Before the advent of practical deep learning, classical search methods enjoyed initial success in AI for games. Many older AI players for board games use minimax search over a game tree, typically augmented with alpha-beta pruning~\citep{alpha-beta-pruning} and game-specific heuristics. A notable early result was DeepBlue~\citep{deepblue}, a 1997 computer chess program based on alpha-beta pruning that defeated then-world champion Garry Kasparov in a six-game match. Even today, the predominant computer chess engine Stockfish uses alpha-beta pruning as its workhorse, decisively achieving superhuman performance on commodity hardware.

However, the game of Go is typically considered to be quite impervious to these classical search techniques, due to the game tree's high branching factor (up to $19 \times 19 + 1 = 362$) and high depth (typically hundreds of moves per game).

\paragraph{Monte Carlo Tree Search (MCTS)}

While some early Go AIs, such as GNUGo~\citep{gnugo}, rely on classical search techniques, most pre-deep learning AIs adopt a technique called Monte Carlo Tree Search \citep[MCTS; ][{}]{mcts-survey}. MCTS treats game tree traversal as an exploitation/exploration tradeoff. At each state, it prioritizes visiting child nodes that provide a high value (estimated utility), or that have not been thoroughly explored. A common exploitation/exploration heuristic is ``upper confidence bounds applied to trees''~\citep[``UCT'';][{}]{uct}; briefly, UCT provides an exploration bonus proportional to the inverse square root of a game state's visit frequency. Go AIs employing MCTS include Leela~\citep{leela}, Pachi~\citep{pachi}, and Fuego~\citep{fuego}.

\paragraph{Early deep learning for Go}

Early attempts at applying deep learning to Go introduced neural networks toward understanding individual game states, usually by predicting win probabilities and best actions from a given state. Go's square grid game board and the spatial locality of moves naturally suggest the use of convolutional architectures, trained on historical human games~\citep{clark2015training,maddison2014move,tian2015darkforest}. AlphaGo~\citep{silver2016alphago} employs policy networks trained with human games and RL, value networks trained via selfplay, and distributed MCTS, achieving a remarkable 4:1 match victory against professional player Lee Sedol in 2016.

\subsection{AlphaGo Zero and AlphaZero}

The AlphaGo Zero~\citep[``AGZ''; ][{}]{silver2017alphagozero} and AlphaZero~\citep[``AZ''; ][{}]{silver2018alphazero} algorithms train a Go AI using no external information except the rules of the game. We provide a high-level overview of AGZ, then briefly describe the similar AZ algorithm.

\paragraph{Move generation algorithm}

The workhorse of AGZ is a residual network model~\citep{resnet}. The model accepts as input a spatially encoded representation of the game state. It then produces a scalar value prediction, representing the probability of the current player winning, and a policy prediction, representing the model's priors on available moves given the current board situation. 

AGZ combines this network with MCTS, which is used as a \emph{policy improvement operator}. Initially informed by the network's current-move policy, the MCTS operator explores the game tree, visiting a new game state at each iteration and evaluating the network policy. It uses a variant of PUCT~\citep{rosin2011multi} to balance exploration (i.e. visiting game states suggested by the prior policy) and exploitation (i.e. visiting game states that have a high value), trading off between the two with a $\cpuct$ constant.

MCTS terminates after a certain number of iterations and produces a new policy based on the visit frequencies of its children in the MCTS game tree. It selects the next move based on this policy, either proportionally (for early-stage training moves), or greedily (for late-stage training moves and all gameplay inference). MCTS can be multithreaded using the virtual loss technique~\citep{virtual-loss}, and MCTS's performance intuitively improves as the number of iterations (``rollouts'') increases.

\paragraph{Training}

AGZ trains the model using randomly sampled data from a replay buffer (filled with selfplay games). The optimization objective is defined as follows, where $V$ and $\vp$ are outputs of a neural network with parameters $\theta$, and $z$ and $\vpi$ are respectively the game outcome and the saved MCTS-augmented policy from the game record:
\begin{equation}
    J(\theta) = |V(s; \theta) - z|^2 - \vpi^T \log \boldsymbol{\vp}(\cdot|\theta) + c\|\theta\|^2 \label{eq:training}
\end{equation}

There are four major components of AGZ's training:
\begin{itemize}
    \item The \textbf{replay buffer} is a fixed-capacity FIFO queue of game records. Each game record consists of the game outcome, the move history, and the MCTS-augmented policies at each move.
    \item The \textbf{selfplay workers} continually take the latest model, play out an AI vs. AI (selfplay) game using the model, and send the game record to the replay buffer.
    \item The \textbf{training worker} continually samples minibatches of moves from the replay buffer and performs stochastic gradient descent (SGD) to fit the model. Every 1,000 minibatches, it sends the model to the evaluator.
    \item The \textbf{evaluator} receives proposed new models. It plays out 400 AI vs. AI games between the new model and the current model and accepts any new model with a 55\% win rate or higher. Accepted models are published to the selfplay workers as the latest model.
\end{itemize}

\paragraph{Performance} Using Tensor Processing Units (TPUs) for selfplays and GPUs for training, AGZ is able to train a 256-filter, 20-block residual network model to superhuman performance in 3 days, and a 256-filter, 40-block model to an estimated 5185 ELO in 40 days. The total computational cost is unknown, however, as the paper does not elaborate on the number of selfplay workers, which we conjecture to be the costliest component of AGZ's training.

\paragraph{AlphaZero} 

While most details of the subsequently proposed AlphaZero (AZ) algorithm~\citep{silver2018alphazero} are similar to those of AGZ, AZ eliminates AGZ's evaluation requirement, thus greatly simplifying the system and allowing new models to be immediately deployed to selfplay workers. AZ provides a remarkable speedup over AGZ, reaching in just eight hours the performance of the aforementioned three-day AGZ model.~\footnote{A caveat here that hardware details are provided for AZ but not AGZ, making it difficult to compare total resource usage.} AZ uses 5,000 first-generation TPU selfplay workers; however, unlike AGZ, AZ uses TPUs for training as well. Beyond Go, the AZ algorithm can be used to train strong chess and shogi AIs.

\subsection{Contemporary implementations}

There are a number of contemporary open-source works that aim to reproduce the performance of AGZ and AZ, also with no external data. LeelaZero~\citep{leelazero} leverages crowd-sourcing to achieve superhuman skill. PhoenixGo~\citep{phoenixgo}, AQ~\citep{aqgo}, and MiniGo~\citep{minigo} achieve similar performance to that of LeelaZero. There are also numerous proprietary implementations, including ``FineArt'', ``Golaxy'', ``DeepZen'', ``Dolbaram'', ``Baduki'', and of course the original implementations of DeepMind~\citep{silver2017alphagozero, silver2018alphazero}.

To our knowledge, \elfopengo{} is the strongest open-source Go AI at the time of writing (under equal hardware constraints), and it has been publicly verified as superhuman via professional evaluation.

\subsection{Understanding AlphaGo Zero and AlphaZero}

The release of AGZ and AZ has motivated a line of preliminary work~\citep{addanki2019understanding,dong2017demystifying,wang2018gameplay,alphax} which aims to analyze and understand the algorithms, as well as to apply similar algorithms to other domains. We offer \elfopengo{} as an accelerator for such research.

\section{\elfopengo{}}

Our proposed \elfopengo{} aims to faithfully reimplement AGZ and AZ, modulo certain ambiguities in the original papers and various innovations that enable our system to operate entirely on commodity hardware. For brevity, we thoroughly discuss the system and software design of \elfopengo{} in \cref{section:thorough-design}; highlights include (1) the colocation of multiple selfplay workers on GPUs to improve throughput, and (2) an asynchronous selfplay workflow to handle the increased per-game latency. Both our training and inference use NVIDIA Tesla V100 GPUs with 16 GB of memory.~\footnote{One can expect comparable performance on most NVIDIA GPUs with Tensor Cores (e.g. the RTX 2060 commodity GPU).}

\paragraph{Comparison of training details}

In \cref{section:thorough-design}'s \cref{table:parameters}, we provide a comprehensive list of hyperparameters, resource provisioning, and other training details of AGZ, AZ, and \elfopengo{}.

To summarize, we largely adhere to AZ's training details. Instead of 5,000 selfplay TPUs and 64 training TPUs, we use 2,000 selfplay GPUs and 8 training GPUs. Since AZ's replay buffer size is unspecified in \citet{silver2018alphazero}, we use the AGZ setting of 500,000 games. We use the AGZ selfplay rollout setting of 1,600 per move. Finally, we use a $\cpuct$ constant of 1.5 and a virtual loss constant of 1.0; these settings are unspecified in \citet{silver2017alphagozero} and \citet{silver2018alphazero}, and we discuss these choices in greater detail in \cref{section:ablation}.

\citet{silver2017alphagozero} establish that during selfplay, the MCTS temperature is set to zero after 30 moves; that is, the policy becomes a Dirac (one-hot) distribution that selects the most visited move from MCTS. However, it is left ambiguous whether a similar temperature decrease is used during training. \elfopengo{}'s training worker uses an MCTS temperature of 1 for all moves (i.e. policy proportional to MCTS visit frequency).

\citet{silver2018alphazero} suggest an alternative, dynamic variant of the PUCT heuristic which we do not explore. We use the same PUCT rule as that of AGZ and the initial December 2017 version of AZ.

\paragraph{Model training specification}

Our main training run constructs a 256-filter, 20-block model (starting from random initialization). First, we run our \elfopengo{} training system for 500,000 minibatches at learning rate $10^{-2}$. Subsequently, we stop and restart the training system twice (at learning rates $10^{-3}$, and $10^{-4}$), each time for an additional 500,000 training minibatches. Thus, the total training process involves 1.5 million training minibatches, or roughly 3 billion game states. We observe a selfplay generation to training minibatch ratio of roughly 13:1 (see \cref{section:thorough-design}). Thus, the selfplay workers collectively generate around 20 million selfplay games in total during training. This training run takes around 16 days of wall clock time, achieving superhuman performance in 9 days.

\paragraph{Practical lessons} We learned a number of practical lessons in the course of training \elfopengo{} (e.g. staleness concerns with typical implementations of batch normalization). For brevity, we relate these lessons in \cref{section:practical-lessons}.

\section{Validating model strength}

We validate \elfopengo{}'s performance via (1) direct evaluation against humans and other AIs, and (2) indirect evaluation using various human and computer-generated game datasets. Unless stated otherwise, \elfopengo{} uses a single NVIDIA Tesla V100 GPU in these validation studies, performing 1,600 rollouts per move.

Based on pairwise model comparisons, we extract our ``final'' model from the main training run after 1.29 million training minibatches. The final model is approximately 150 ELO stronger than the prototype model, which we immediately introduce. Both the prototype and final model are publicly available as pretrained models.

\subsection{Prototype model}

Human benchmarking is essential for determining the strength of a model. For this purpose, we use an early model trained in April 2018 and subsequently evaluated against human professional players. We trained this 224-filter, 20-block model using an experimental hybrid of the AGZ and AZ training processes. We refrain from providing additional details on the model, since the code was modified during training resulting in the model not being reproducible. However, it provides a valuable benchmark. We refer to this model as the ``prototype model''.

We evaluate our prototype model against 4 top 30 professional players.~\footnote{The players (and global ranks as of game date) include Kim Ji-seok (\#3), Shin Jin-seo (\#5), Park Yeonghun (\#23), and Choi Cheolhan (\#30). All four players were fairly compensated for their expertise, with additional and significant incentives for winning versus \elfopengo{}. Each player played 5 games, for a total of 20 human evaluation games.} \elfopengo{} plays under 50 seconds per move ($\approx$ 80,000 rollouts), with no pondering during the opponent's turn, while the humans play under no time limit. These evaluation games typically last for 3-4 hours, with the longest game lasting over 6 hours. Using the prototype model, \elfopengo{} won every game for a final record of 20:0.

\begin{figure}[t]
\centering
\includegraphics[draft=false, width=0.99\linewidth]{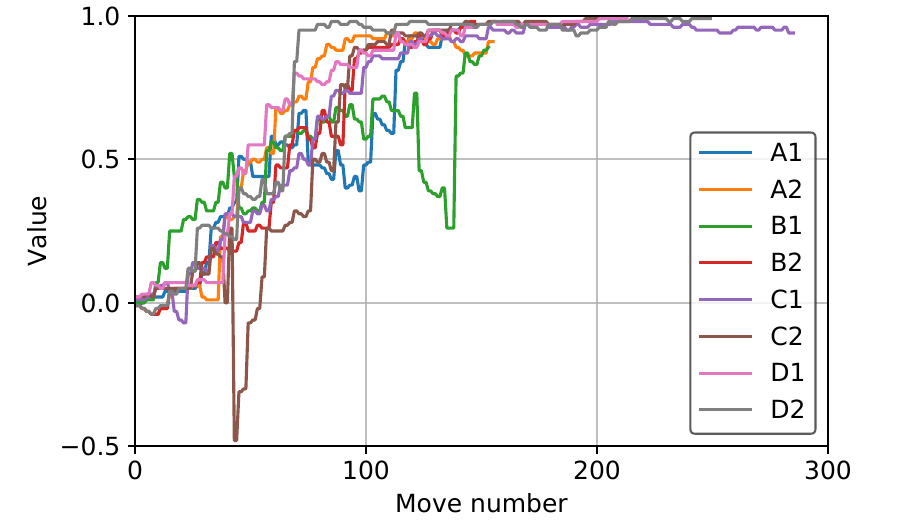}
\caption{Predicted values after each move for 8 selected games versus human professional players (2 games per player). A positive value indicates that the model believes it is more likely to win than the human. Per players' request, we anonymize the games by arbitrarily labeling the players with letters `A' to `D'.}
\label{fig:predicted-value}
\end{figure}

\cref{fig:predicted-value} depicts the model's predicted value during eight selected human games; this value indicates the perceived advantage of \elfopengo{} versus the professionals  over the course of the game. Sudden drops of the predicted value indicate that \elfopengo{} has various weaknesses (e.g. ladder exploitation); however, the human players ultimately proved unable to maintain the consequent advantages for the remainder of the game.

\paragraph{Evaluation versus other AIs}

We further evaluate our prototype model against LeelaZero~\citep{leelazero}, which at the time of evaluation was the strongest open-source Go AI.~\footnote{2018 April 25 model with 192 filters and 15 residual blocks; public hash \texttt{158603eb}.} Both LeelaZero and \elfopengo{} play under a time limit of 50 seconds per move. \elfopengo{} achieves an overall record of 980:18, corresponding to a win rate of 98.2\% and a skill differential of approximately 700 ELO.

\subsection{Training analysis} \label{section:training-analysis}

Throughout our main training run, we measure the progression of our model against the human-validated prototype model. We also measure the agreement of the model's move predictions with those made by humans. Finally, we explore the rate at which the model learns different stages of the game and complex moves, such as ladders.

\paragraph{Training progression: selfplay metrics}

\begin{figure}[t]
\centering
\includegraphics[draft=false, width=0.99\linewidth]{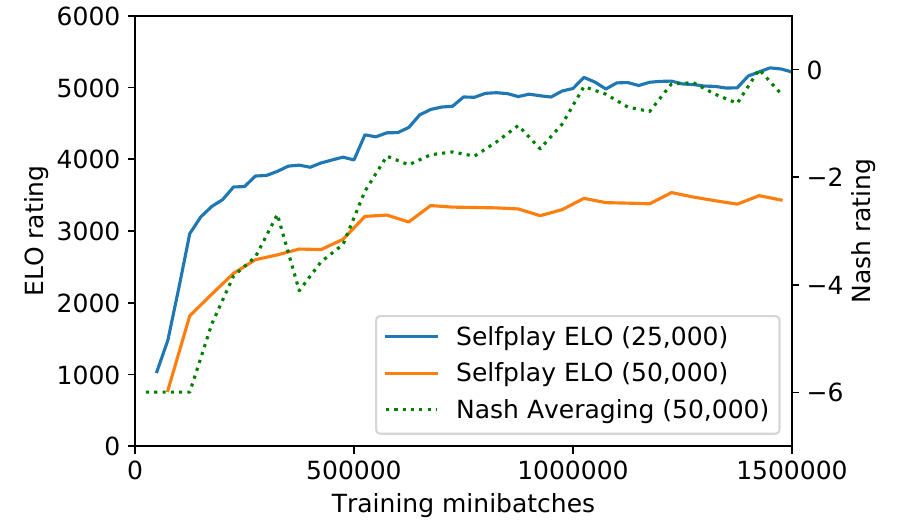}
\caption{Progression of model skill during training. ``Selfplay ELO 25,000'' and ``Selfplay ELO 50,000'' refer to the unnormalized selfplay ELO rating, calculated based on consecutive model pairs at intervals of 25,000 and 50,000 training minibatches, respectively. ``Nash Averaging 50,000'' refers to the Nash averaging rating~\citep{nash-eval}, calculated based on a pairwise tournament among the same models as in ``Selfplay ELO 50,000''.}
\label{fig:elo-rating}
\end{figure}

In examining the progression of model training, we consider two selfplay rating metrics: selfplay ELO and Nash averaging. Selfplay ELO uses \citet{elo}'s rating formula in order to determine the rating difference between consecutive pairs of models, where each pair's winrate is known. Nash averaging~\citep{nash-eval} calculates a logit-based payoff of each model against a mixed Nash equilibrium, represented as a discrete probability distribution over each model.

\begin{figure*}[!t]
\centering
\begin{subfigure}[b]{0.33\linewidth}
\includegraphics[draft=false, width=1.0\linewidth]{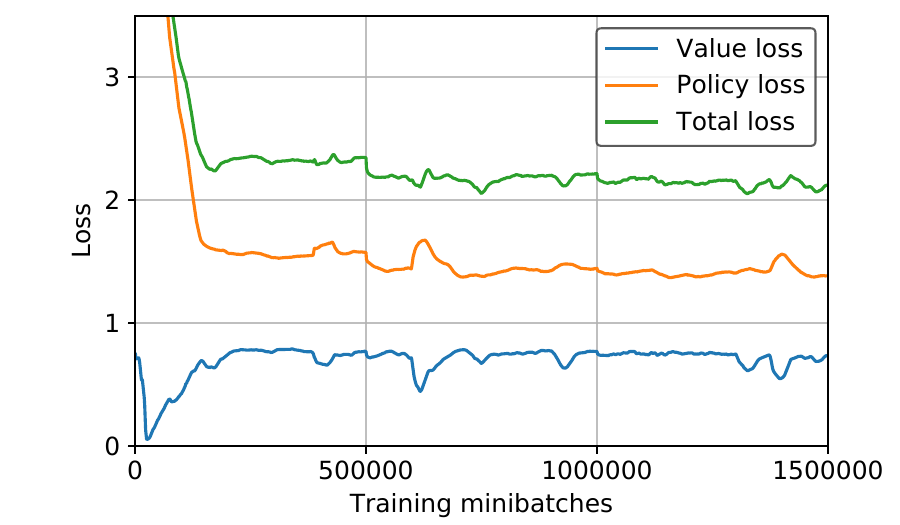}
\caption{} \label{fig:training_losses}
\end{subfigure}
\begin{subfigure}[b]{0.33\linewidth}
\includegraphics[draft=false, width=1.0\linewidth]{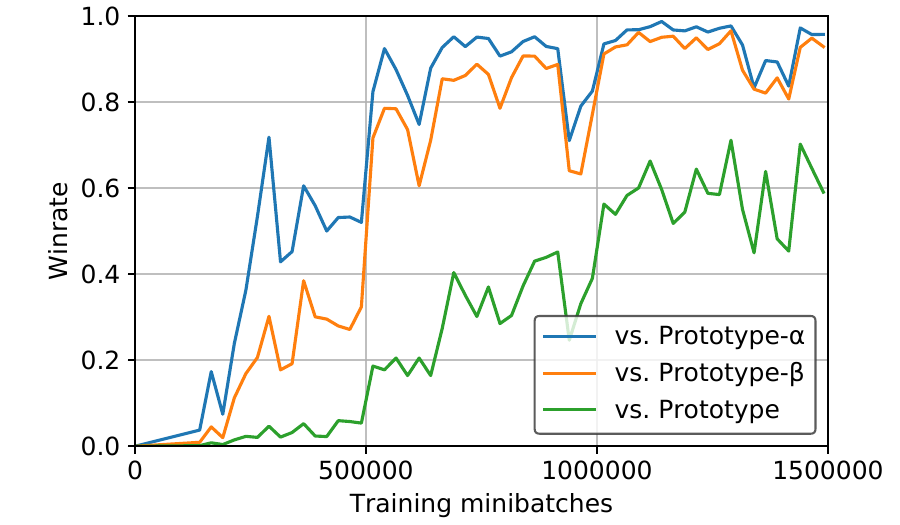}
\caption{} \label{fig:mainline_prototype}
\end{subfigure}
\begin{subfigure}[b]{0.33\linewidth}
\includegraphics[draft=false, width=1.0\linewidth]{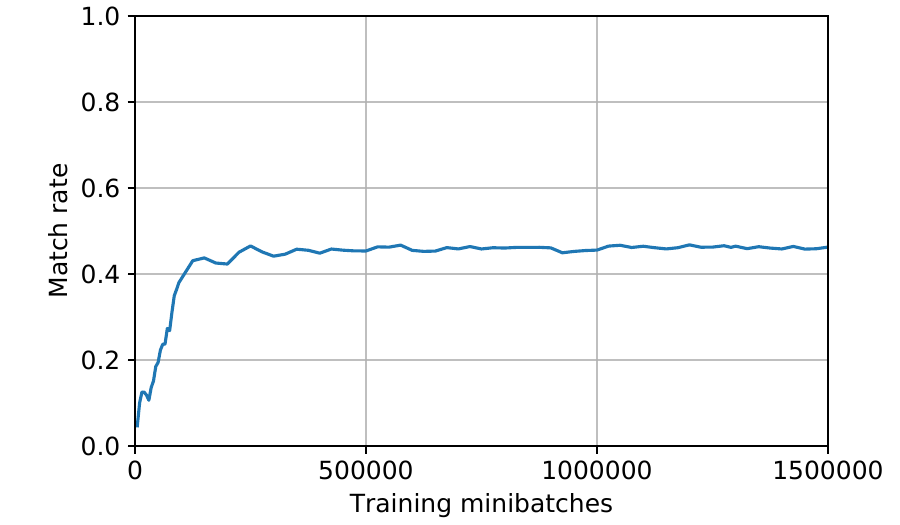}
\caption{} \label{fig:match_full}
\end{subfigure}
\caption{(a) Model's training loss (value, policy, and sum) during training. (b) Win rates vs. the prototype model during training. (c) Match rate between trained model and human players on moves from professional games (as described in \cref{section:human-dataset-details}). The learning rate was decreased every 500,000 minibatches.}
\label{fig:training-analysis}
\end{figure*}

Intuitively, the selfplay ELO can be viewed as an ``inflated'' metric for two main reasons. First, the model's rating will increase as long as it can beat the immediately preceding model, without regard to its performance against earlier models. Second, the rating is sensitive to the number of consecutive pairs compared; increasing this will tend to boost the rating of each model due to the nonlinear logit form of the ELO calculation.

\cref{fig:elo-rating} shows the selfplay ELO rating using every 25,000-th model and every 50,000-th model, and the Nash averaging rating using every 50,000-th model. The ratings are consistent with the above intuitions; the ELO ratings follow a mostly consistent upward trend, and the denser comparisons lead to a more inflated rating. Note that the Nash averaging rating captures model skill degradations (particularly between minibatches 300,000 and 400,000) that selfplay ELO fails to identify.

\paragraph{Training progression: training objective}

\cref{fig:training_losses} shows the progression of the policy and value losses. Note the initial dip in the value loss. This is due to the model overestimating the white win rate, causing the black player to resign prematurely, which reduces the diversity of the games in the replay buffer. This could result in a negative feedback loop and overfitting. \elfopengo{} automatically corrects for this by evenly sampling black-win and white-win games. With this diverse (qualitatively, ``healthy'') set of replays, the value loss recovers and stays constant throughout the training, showing there is always new information to learn from the replay buffer.

\paragraph{Performance versus prototype model}

\cref{fig:mainline_prototype} shows the progression of the model's win rates against two weaker variants taken earlier in training of the prototype model  (``prototype-$\alpha$'' and ``prototype-$\beta$'') as well as the main human-validated prototype model (simply ``prototype'').~\footnote{Prototype-$\alpha$ is roughly at the advanced amateur level, and prototype-$\beta$ is roughly at a typical professional level.} We observe that the model trends stronger as training progresses, achieving parity with the prototype after roughly 1 million minibatches, and achieving a 60\% win rate against the prototype at the end of training. Similar trends emerge with prototype-$\alpha$ and prototype-$\beta$, demonstrating the robustness of the trend to choice of opponent. Note that while the strength of the model trends stronger, there is significant variance in the model's strength as training progresses, even with weaker models. Surprisingly, this variance does not decrease even with a decrease in learning rate. 

Examining the win rates versus other models in \cref{fig:mainline_prototype}, the number of minibatches could potentially be reduced to 250,000 at each learning rate, and still similar performance can be achieved.

\paragraph{Comparison with human games}
Since our model is significantly stronger than the prototype model that showed superhuman strength, we hypothesize that our model is also of superhuman strength. In \cref{fig:match_full}, we show the agreement of predicted moves with those made by professional human players. The moves are extracted from 1,000 professional games played from 2011 to 2015. The model quickly converges to a human move match rate of $\approx$ 46\% around minibatch 125,000. This indicates that the strengthening of the model beyond this point may not be due to better human professional predictions, and as demonstrated by~\citet{silver2016alphago}, there may be limitations to supervised training from human games.

\paragraph{Learning game stages}

An interesting question is whether the model learns different stages of the game at different rates. During training, is the model initially stronger at opening or endgame moves?

We hypothesize that the endgame should be learned earlier than the opening. With a random model, MCTS behaves randomly except for the last few endgame moves, in which the actual game score is used to determine the winner of the game. Then, after some initial training, the learned endgame signals inform MCTS in the immediately preceding moves as well. As training progresses, these signals flow to earlier and earlier move numbers via MCTS, until finally the opening is learned.

\begin{figure*}[!t]
\centering
\begin{subfigure}[b]{0.33\linewidth}
\includegraphics[draft=false, width=1.0\linewidth]{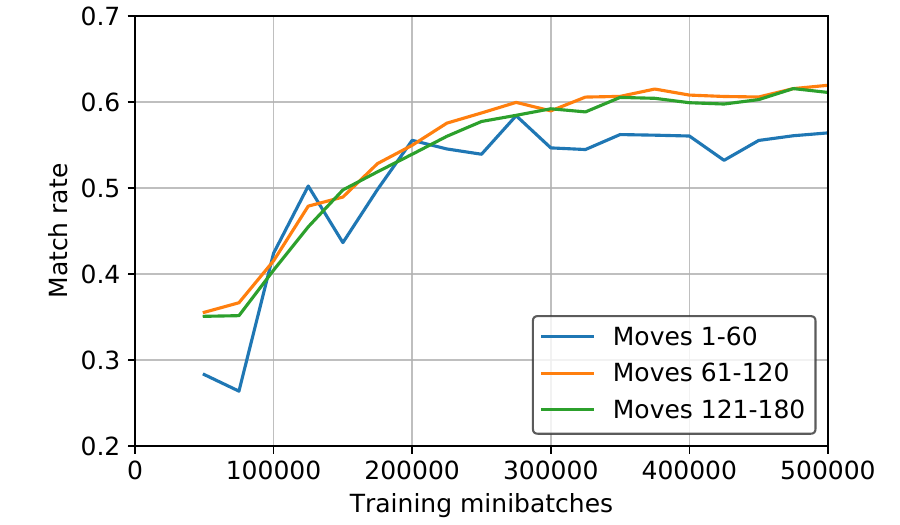}
\caption{} \label{fig:selfplay_match}
\end{subfigure}
\begin{subfigure}[b]{0.33\linewidth}
\includegraphics[draft=false, width=1.0\linewidth]{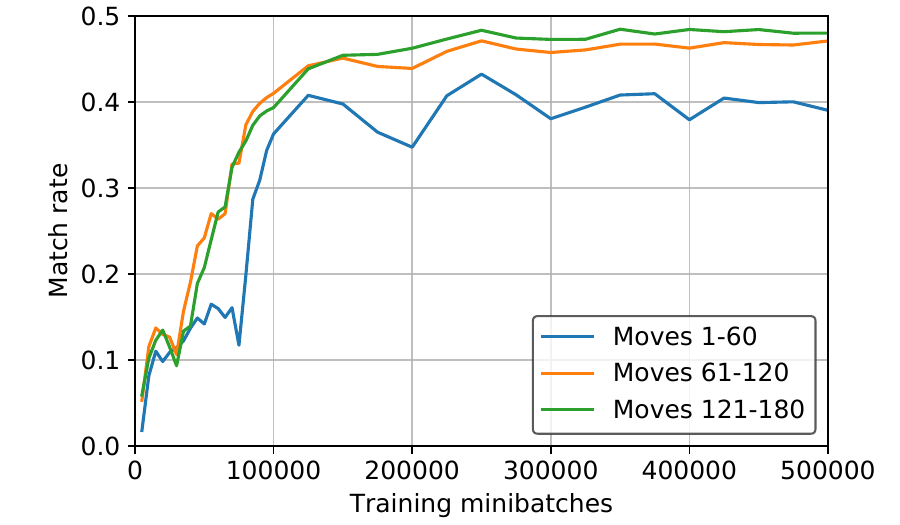}
\caption{} \label{fig:match}
\end{subfigure}
\begin{subfigure}[b]{0.33\linewidth}
\includegraphics[draft=false, width=1.0\linewidth]{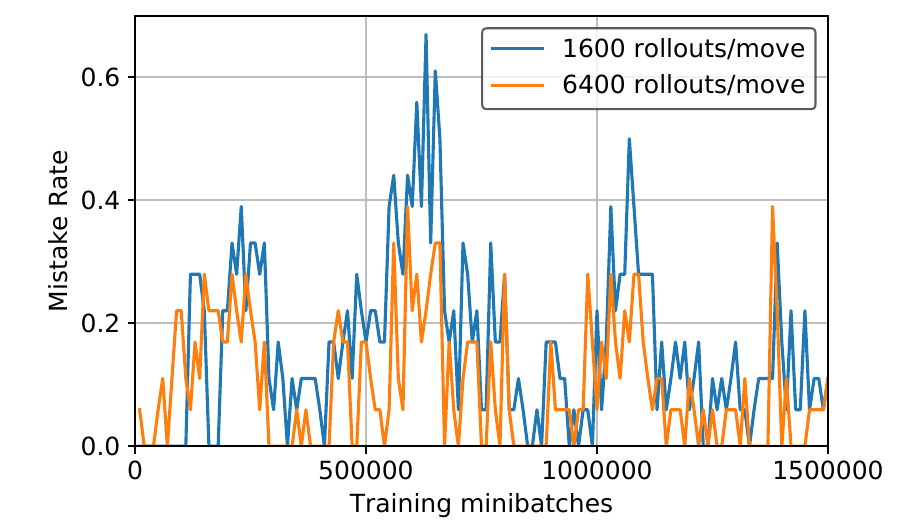}
\caption{} \label{fig:ladder_mistake}
\end{subfigure}
\caption{(a) Move agreement rates with the prototype model at different stages of the game during training. Note that the model learns better end-game moves before learning opening moves. The first part of the figure is clipped due to short-game dominance cased by initial overfitting issues. (b) Move agreement rates with human games at different stages of the game during the first stage of training. (c) Model's rate of ``mistake moves'' on the ladder scenario dataset during training, associated with vulnerability to ladder scenarios.}
\end{figure*}

In \cref{fig:selfplay_match} and \cref{fig:match}, we show the agreement of predicted moves from the model during training, and the prototype model and humans respectively. \cref{fig:selfplay_match} shows the percentage of moves at three stages of gameplay (moves 1-60, 61-120, and 121-180) from training selfplay games that agree with those predicted by the prototype model. \cref{fig:match} is similar, but it uses moves from professional human games, and measures the match rate between the trained model and professional players. Consistent with our hypothesis, the progression of early games moves (1-60) lags behind that of the mid-game moves (61-120) and the end-game moves (121-180). Upon further exploration, we observed that this lag resulted from some initial overfitting before eventually recovering. Counter to conventional wisdom, the progression of match rates of mid-game and end-game moves are nearly identical. Note that more ambiguity exists for earlier moves than later moves, so after sufficient training the match rates converge to different match rates.

\paragraph{Ladder moves}

\begin{figure}[h]
\centering
\includegraphics[draft=false, width=0.48\linewidth]{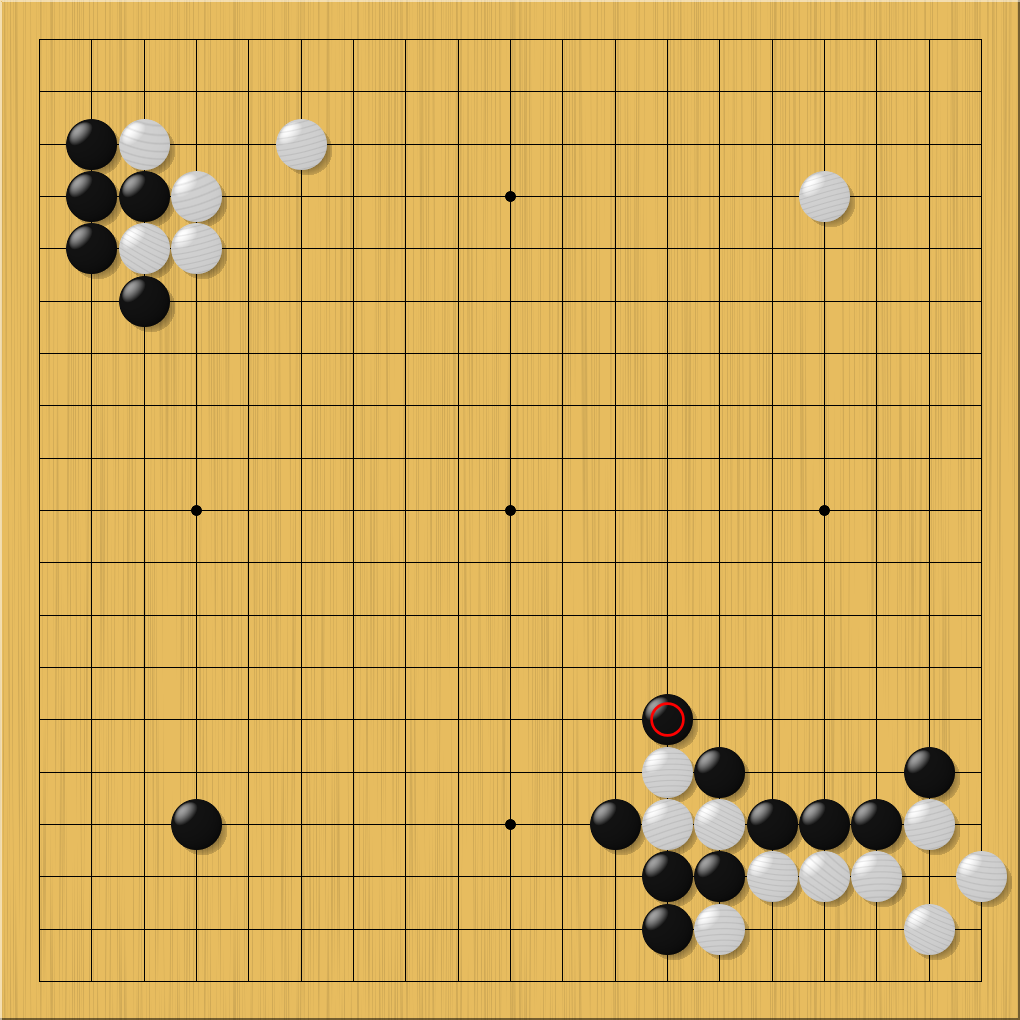}
\hfill
\includegraphics[draft=false, width=0.48\linewidth]{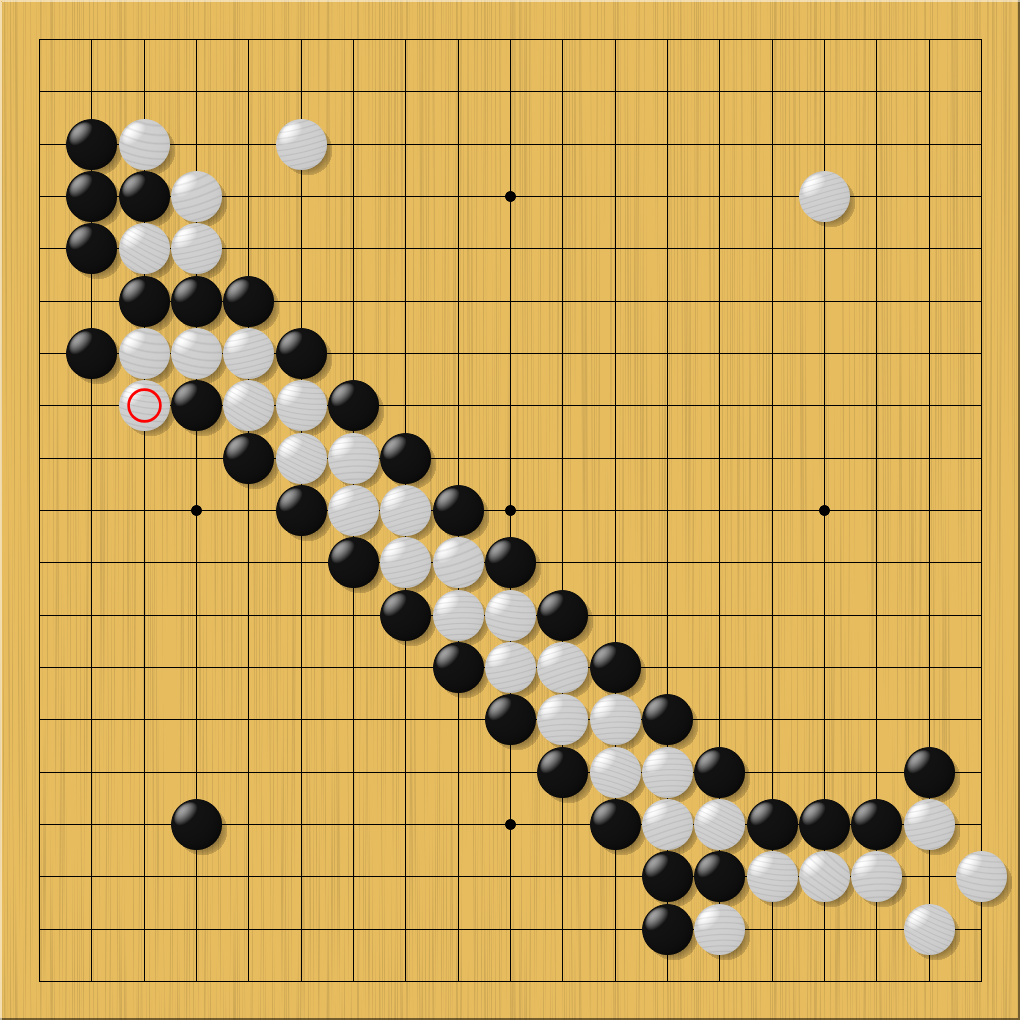}
\caption{The beginning (left) and end (right) of a ladder scenario in which OpenGo (black) mistakenly thought it could gain an advantage by playing the ladder. Whether a ladder is advantageous or not is commonly dependent on stones distant from its start.}
\label{fig:ladder-example}
\end{figure}

``Ladder'' scenarios are among the earliest concepts of Go learned by beginning human players. Ladders create a predictable sequence of moves for each player that can span the entire board as shown in \cref{fig:ladder-example}. In contrast to humans, \citet{silver2017alphagozero} observes that deep RL models learn these tactics very late in training. To gain further insight into this phenomenon, we curate a dataset of 100 ladder scenarios (as described in \cref{section:ladder-dataset-details}) and evaluate the model's ladder handling abilities throughout training.

As shown in \cref{fig:ladder_mistake}, we observe that the ability of the network to correctly handle ladder moves fluctuates significantly over the course of training. In general, ladder play improves with a fixed learning rate and degrades after the learning rate is reduced before once again recovering. While the network improves on ladder moves, it still makes mistakes even late in training. In general, increasing the number of MCTS rollouts from 1,600 to 6,400 improves ladder play, but mistakes are still made.

\section{Ablation study} \label{section:ablation}

We employ \elfopengo{} to perform various ablation experiments, toward demystifying the inner workings of MCTS and AZ-style training.

\subsection{PUCT constant}

\begin{figure}[h]
\centering
\includegraphics[draft=false, width=0.99\linewidth]{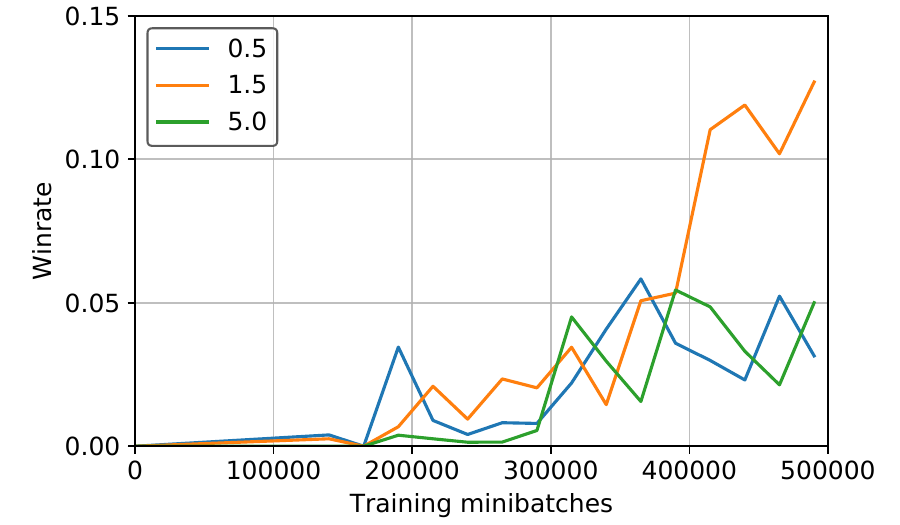}
\caption{Win rates vs. the prototype model during an abbreviated training run for three different $\cpuct$ values. The learning rate is decayed 10-fold after 290,000 minibatches.}
\label{fig:puct-sweep}
\end{figure}

Both AGZ \citep{silver2017alphagozero} and AZ \citep{silver2018alphazero} leave $\cpuct$ unspecified. Recall that $\cpuct$ controls the balance of exploration vs. exploitation in the MCTS algorithm. Thus, it is a crucial hyperparameter for the overall behavior and effectiveness of the algorithm. Setting $\cpuct$ too low results in insufficient exploration, while setting $\cpuct$ too high reduces the effective depth (i.e. long-term planning capacity) of the MCTS algorithm. In preliminary experimentation, we found $\cpuct = 1.5$ to be a suitable choice. \cref{fig:puct-sweep} depicts the \elfopengo{} model's training trajectory under various values of $\cpuct$; among the tested values, $\cpuct = 1.5$ is plainly the most performant.

\subsection{MCTS virtual loss}

\begin{table}[ht]
    \centering
    \begin{tabular}{|c|c|c|c|c|c|c|}
      \hline
      0.1 & 0.2 & 0.5 & 1.0 & 2.0 & 3.0 & 5.0 \\ \hline
      22\% & 37\% & 49\% & 50\% & 32\% & 36\% & 30\% \\ \hline
    \end{tabular}
    \caption{Win rates of various virtual loss constants versus the default setting of 1.}
    \label{table:vloss}
\end{table}

Virtual loss~\citep{virtual-loss} is a technique used to accelerate multithreaded MCTS. It adds a temporary and fixed amount of loss to a node to be visited by a thread, to prevent other threads from concurrently visiting the same node and causing congestion. The virtual loss is parameterized by a constant. For AGZ, AZ, and \elfopengo{}'s value function, a virtual loss constant of 1 is intuitively interpretable as each thread temporarily assuming a loss for the moves along the current rollout. This motivates our choice of 1 as \elfopengo{}'s virtual loss constant.

We perform a sweep over the virtual loss constant using the prototype model, comparing each setting against the default setting of 1. The results, presented in \cref{table:vloss}, suggest that a virtual loss constant of 1 is indeed reasonable.

\subsection{AlphaGo Zero vs. AlphaZero}

We hypothesize that AZ training vastly outperforms AGZ training (given equivalent hardware) due to the former's asynchrony and consequent improved throughput. We train 64-filter, 5-block models from scratch with both AGZ and AZ for 12 hours and compare their final performance. The model trained with AZ wins 100:0 versus the model trained with AGZ, indicating that AZ is indeed much more performant.

\subsection{MCTS rollout count}

\begin{figure}[t]
\centering
\includegraphics[draft=false,width=0.99\linewidth]{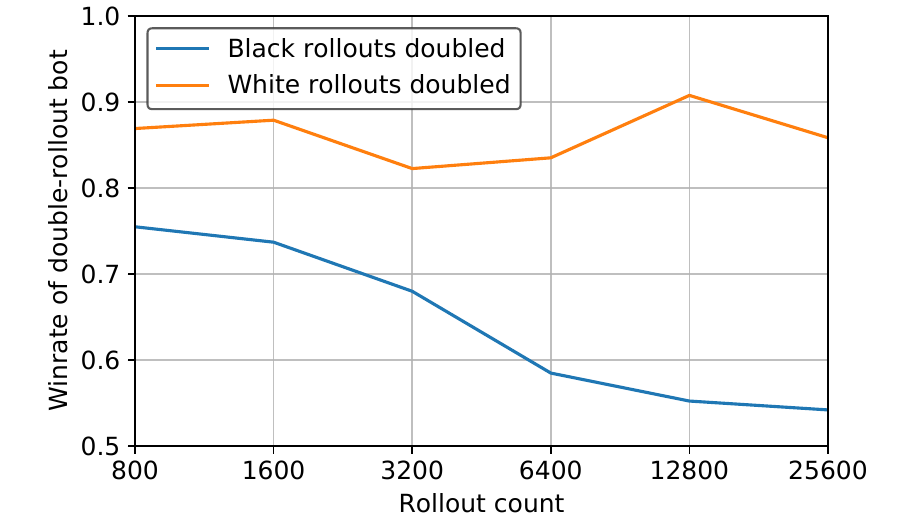}
\caption{Win rate when playing with 2x rollouts against the same model with 1x rollouts.}
\label{fig:rollout-sweep-bs16}
\end{figure}

Intuitively, increasing the number of MCTS iterations (rollout count) improves the AI's strength by exploring more of the game tree. Toward better understanding the rollout count's impact on strength, we perform a selfplay analysis with the final model, in which one player uses twice as many MCTS rollouts as the other. We perform this analysis across a wide range of rollout counts (800-25,600).

From the results shown in \cref{fig:rollout-sweep-bs16}, we find that \elfopengo{} consistently enjoys an 80-90\% win rate ($\approx$ 250-400 additional ELO) from doubling the number of rollouts as the white player. On the other hand, as the black player, \elfopengo{} only enjoys a 55-75\% win rate ($\approx$ 35-200 additional ELO). Moreover, the incremental benefit for black of doubling rollouts shrinks to nearly 50\% as the rollout count is increased, suggesting that our model has a skill ceiling with respect to rollout count as the black player. That this skill ceiling is not present as the white player suggests that a 7.5 komi (white score bonus) can be quite significant for black.

Since using the same model on both sides could introduce bias (the player with more rollouts sees all the branches explored by the opponent), we also experiment with the prototype/final model and still observe a similar trend that doubling the rollouts gives $\approx$ 200 ELO boost.

\section{Discussion}

\elfopengo{} learns to play the game of Go differently from humans. It requires orders of magnitude more games than professional human players to achieve the same level of performance. Notably, ladder moves, which are easy for human beginners to understand after a few examples, are difficult for the model to learn and never fully mastered. We suspect that this is because convolutional networks lack the right inductive bias to handle ladders and resort to memorizing specific situations. Finally, we observe significant variance during training, which indicates the method is still not fully understood and may require much tuning. We hope this paper serves as a starting point for improving upon AGZ/AZ to achieve efficient and stable learning.  

Surprisingly, further reduction of the learning rate does not improve training. We trained to 2 million minibatches with a learning rate of $10^{-5}$, but noticed minimal improvement in model strength. Furthermore, the training becomes unstable with high variance. This may be due to a lack of diversity in the selfplay games, since either the model has saturated or the lower learning rate of $10^{-5}$ resulted in the selfplay games coming from nearly identical models. It may be necessary to increase the selfplay game buffer when using lower learning rates to maintain stability.

Finally, RL methods typically learn from the states that are close to the terminal state (end game) where there is a sparse reward. Knowledge from the reward is then propagated towards the beginning of the game. However, from \cref{fig:selfplay_match} and \cref{fig:match}, such a trend is weak (moves 61-180 are learned only slightly faster than moves 1-60). This brings about the question of why AGZ/AZ methods behave differently, and what kind of role the inductive bias of the model plays during the training. It is likely that due to the inductive bias, the model quickly predicts the correct moves and values of easy situations, and then focuses on the difficult cases.

\section{Conclusion}

We provide a reimplementation of AlphaZero~\citep{silver2018alphazero} and a resulting Go engine capable of superhuman gameplay. Our code, models, selfplay datasets and auxiliary data are publicly available. We offer insights into the model's behavior during training. Notably, we examine the variance of the model's strength, its ability to learn ladder moves, and the rate of improvement at different stages of the game. Finally, through a series of ablation studies, we shed light on parameters that were previously ambiguous. Interestingly, we observe significant and sustained improvement with the number of rollouts performed during MCTS when playing white, but diminishing returns when playing black. This indicates that the model's strength could still be significantly improved. 

Our goal is to provide the insights, code, and datasets necessary for the research community to explore large-scale deep reinforcement learning. As demonstrated through our experiments, exciting opportunities lie ahead in exploring sample efficiency, reducing training volatility, and numerous additional directions.


\bibliography{main}
\bibliographystyle{icml2019}

\clearpage
\appendix

\newcommand{\beginsupplement}{%
    \setcounter{table}{0}
    \renewcommand{\thetable}{S\arabic{table}}%
    \setcounter{figure}{0}
    \renewcommand{\thefigure}{S\arabic{figure}}%
}

\beginsupplement

\def\alphazero{AlphaZero}
\def\alphagozero{AlphaGo Zero}

\begin{table*}[!t]
    \centering
    \begin{tabular}{|c||c|c|c|}
    \hline
    \thead{Parameter/detail} & \thead{AGZ} & \thead{AZ} & \thead{\elfopengo{}} \\
    \hline
    $\cpuct$ (PUCT constant) & ? & ? & 1.5 \\ \hline
    MCTS virtual loss constant & ? & ? & 1.0 \\ \hline
    MCTS rollouts (selfplay) & 1,600 & 800 & 1,600 \\ \hline
    Training algorithm & \multicolumn{3}{c|}{SGD with momentum = 0.9} \\ \hline
    Training objective & \multicolumn{3}{c|}{$\text{value squared error} + \text{policy cross entropy} + 10^{-4} \cdot \text{L2}$} \\ \hline
    Learning rate & $10^{\{-2, -3, -4\}}$ & $2 \cdot 10^{\{-2, -3, -4\}}$ & $10^{\{-2, -3, -4\}}$ \\ \hline
    Replay buffer size & 500,000 & ? & 500,000 \\ \hline
    Training minibatch size & 2048 & 4096 & 2048 \\ \hline
    Selfplay hardware & ? & 5,000 TPUs & 2,000 GPUs \\ \hline
    Training hardware & 64 GPUs & 64 TPUs & 8 GPUs \\ \hline
    Evaluation criteria & 55\% win rate & none & none \\ \hline
    \end{tabular}
    \caption{Hyperparameters and training details of AGZ, AZ, and \elfopengo{}. ``?'' denotes a detail that was ambiguous or unspecified in \citet{silver2017alphagozero} or \citet{silver2018alphazero}}
    \label{table:parameters}
\end{table*}

\section{Detailed system and software design of \elfopengo{}} \label{section:thorough-design}

We now describe the system and software design of \elfopengo{}, which builds upon the DarkForest Go engine~\citep{tian2015darkforest} and the ELF reinforcement learning platform~\citep{tian2017elf}.

\subsection{Distributed platform}

\elfopengo{} is backed by a modified version of ELF~\citep{tian2017elf}, an Extensive, Lightweight and Flexible platform for reinforcement learning research; we refer to the new system as \system{}. Notable improvements include:
\begin{itemize}
    \item \textbf{Distributed support.} \system{} adds support for distributed asynchronous workflows, supporting up to 2,000 clients.
    \item \textbf{Increased Python flexibility.} We provide a distributed adapter that can connect any environment with a Python API, such as OpenAI Gym~\citep{brockman2016openai}.
    \item \textbf{Batching support.} We provide a simple autobatching registry to facilitate shared neural network inference across multiple game simulators.
\end{itemize}

\subsection{Go engine}

We have integrated the DarkForest Go engine~\citep{tian2015darkforest} into \elfopengo{}, which provides efficient handling of game dynamics (approximately $1.2~\mu s$ per move). We use the engine to execute game logic and to score the terminal game state using Tromp-Taylor rules~\citep{tt-scoring}.

\subsection{Training system}

\elfopengo{} largely replicates the training system architecture of \alphazero. However, there are a number of differences motivated by our compute capabilities.

\paragraph{Hardware details}

We use NVIDIA V100 GPUs for our selfplay workers. Every group of eight GPUs shares two Intel Xeon E5-2686v4 processors.

We use a single training worker machine powered by eight V100 GPUs. Increasing the training throughput via distributed training does not yield considerable benefit for \elfopengo{}, as our selfplay throughput is much less than that of \alphazero{}.

To elaborate on this point, our ratio of selfplay games to training minibatches with a single training worker is roughly 13:1. For comparison, \alphazero{}'s ratio is 30:1 and \alphagozero{}'s ratio is 7:1. We found that decreasing this ratio significantly below 10:1 hinders training (likely due to severe overfitting). Thus, since adding more training workers proportionally decreases this ratio, we refrain from multi-worker training.

\paragraph{GPU colocation of selfplay workers}

We use GPUs instead of TPUs to evaluate the residual network model. The primary difference between the two is that GPUs are much slower for residual networks.~\footnote{According to December 2018 DawnBench~\citep{coleman2017dawnbench} results available at \url{https://dawn.cs.stanford.edu/benchmark/ImageNet/train.html}, one TPU has near-equivalent 50-block throughput to that of eight V100 GPUs} Neural networks on GPUs also benefit from batching the inputs; in our case, we observed near-linear throughput improvements from increasing the batch size up to 16, and sublinear but still significant improvements between 16 and 128.

Thus, to close the gap between GPU and TPU, we co-locate 32 selfplay workers on each GPU, allowing the GPU to process inputs from multiple workers in a single batch. Since each worker has 8 game threads, this implies a theoretical maximum of 256 evaluations per batch. In practice, we limit the batch size to 128.

This design, along with the use of half-precision floating point computation, increases the throughput of each \elfopengo{} GPU selfplay worker to roughly half the throughput of a AlphaGo Zero TPU selfplay worker. While \alphagozero{} reports throughput of 2.5 moves per second for a 256-filter, 20-block model with 1,600 MCTS rollouts, \elfopengo{}'s throughput is roughly 1 move per second.

\paragraph{Asynchronous, heterogenous-model selfplays}

This colocation, along with the inherent slowness of GPUs relative to TPUs, results in much higher latency for game generation (on the order of an hour). Since our training worker typically produces a new model (i.e. processes 1,000 minibatches) every 10 to 15 minutes, new models are published faster than a single selfplay can be produced.

There are two approaches to handling this:

\paragraph{Synchronous (AlphaGo Zero) mode} In this mode, there are two different kinds of clients: \texttt{Selfplay} clients and \texttt{Eval} clients. Once the server has a new model, all the \texttt{Selfplay} clients discard the current game being played, reload the new model and restart selfplays, until a given number of selfplays have been generated. Then the server starts to update the current model according to Equation~\ref{eq:training}. Every 1,000 minibatches, the server updates the model and notifies \texttt{Eval} clients to compare the new model with the old one. If the new model is better than the current one by 55\%, then the server notifies all the clients to discard current games, and restart the loop. On the server side, the selfplay games from the previous model can either be removed from the replay buffer or be retained. Otherwise, the \texttt{Selfplay} clients send more selfplays of the current model until an additional number of selfplays are collected by the server, and then the server starts another 1,000 batches of training. This procedure repeats until a new model passes the win rate threshold. 

\paragraph{Asynchronous (AlphaZero) mode} Note that AlphaGo Zero mode involves a lot of synchronization and is not efficient in terms of boosting the strength of the trained model. In AlphaZero mode, we release all the synchronization locks and remove \texttt{Eval} clients. Moves are always generated using the latest models and \texttt{Selfplay} clients do not terminate their current games upon receiving new models. It is possible that for a given selfplay game record, the first part of the game is played by model A while the second part of the game is played by model B.

We initially started with the synchronous approach before switching to the asynchronous approach. Switching offered two benefits: (1) Both selfplay generation and training realized a drastic speedup. The asynchronous approach achieves over 5x the selfplay throughput of the synchronous approach on our hardware setup. (2) The ratio of selfplay games to training minibatches increased by roughly 1.5x, thus helping to prevent overfitting.

The downside of the asynchronous approach is losing homogeneity of selfplays -- each selfplay is now the product of many consecutive models, reducing the internal coherency of the moves. However, we note that the replay buffer that provides training data is already extremely heterogeneous, typically containing games from over 25 different models. Consequently, we suspect and have empirically verified that the effect of within-selfplay heterogeneity is mild.

\subsection{Miscellany}

\paragraph{Replay buffer} On the server side, we use a large replay buffer (500,000 games) to collect game records by clients. Consistent with~\citet{silver2017alphagozero}, who also use a replay buffer of 500,000 games, we found that a large replay buffer yields good performance. To increase concurrency (reading from multiple threads of feature extraction), the replay buffer is split into 50 queues, each with a maximal size of 10,000 games and a minimal size of 200 games. Note that the minimal size is important, otherwise the model often starts training on a very small set of games and quickly overfits before more games arrive.  

\paragraph{Fairness of model evaluation} In synchronous mode, the server deals with various issues (e.g., clients die or taking too long to evaluate) and makes sure evaluations are done in an unbiased manner. Note that a typically biased estimation is to send 1,000 requests to \texttt{Eval} clients and conclusively calculate the win rate using the first 400 finished games. This biases the metric toward shorter games, to training's detriment.

\paragraph{Game resignation} Resigning from a hopeless game is very important in the training process. This not only saves much computation but also shifts the selfplay distribution so that the model focuses more on the midgame and the opening after learning the basics of Go. As such, the model uses the bulk of its capacity for the most critical parts of the game, thus becoming stronger. As in \citet{silver2017alphagozero}, we dynamically calibrate our resignation threshold to have a 5\% false positive rate; we employ a simple sliding window quantile tracker.

\section{Auxiliary dataset details} \label{section:dataset-details}

\subsection{Human games dataset} \label{section:human-dataset-details}

To construct the human game dataset, we randomly sample 1,000 professional games from the Gogod database from 2011 to 2015.~\footnote{\url{https://gogodonline.co.uk/}}

\subsection{Ladder dataset} \label{section:ladder-dataset-details}

We collect 100 games containing ladder scenarios from the online CGOS (Computer Go Server) service, where we deployed our prototype model.~\footnote{\url{http://www.yss-aya.com/cgos/19x19/standings.html}} For each game, we extract the decisive game state related to the ladder. We then augment the dataset 8-fold via rotations and reflections.

\section{Practical lessons} \label{section:practical-lessons}

\elfopengo{} was developed through much iteration and bug-fixing on both the systems and algorithm/modeling side. Here, we relate some interesting findings and lessons learned from developing and training the AI.

\paragraph{Batch normalization moment staleness} Our residual network model, like that of AGZ and AZ, uses batch normalization~\citep{ioffe2015batchnorm}. Most practical implementations of batch normalization use an exponentially weighted buffer, parameterized by a ``momentum constant'', to track the per-channel moments. We found that even with relatively low values of the momentum constant, the buffers would often be stale (biased), resulting in subpar performance.

Thus, we adopt a variant of the postprocessing moment calculation scheme originally suggested by \citet{ioffe2015batchnorm}. Specifically, after every 1,000 training minibatches, we evaluate the model on 50 minibatches and store the simple average of the activation moments in the batch normalization layers. This eliminates the bias in the moment estimators, resulting in noticeably improved and consistent performance during training. We have added support for this technique to the PyTorch framework~\citep{paszke2017pytorch}.~\footnote{As of December 2018, this is configurable by setting \texttt{momentum=None} in the \texttt{BatchNorm} layer constructor.}

\paragraph{Dominating value gradients}

We performed an unintentional ablation study in which we set the cross entropy coefficient to $\frac{1}{362}$ during backpropogation. This change results in optimizing the value network much faster than the policy network. We observe that with this modification, \elfopengo{} can still achieve a strength of around amateur dan level. Further progress is extremely slow, likely due to the minimal gradient provided by the policy network. This suggests that any MCTS augmented with only a value heuristic has a relatively low skill ceiling in Go.

\end{document}